\documentclass{article}

\usepackage[dblblindworkshop, final]{neurips_2026}
\workshoptitle{Linguistic Principles for Foundation Models (LP4FM)}

\usepackage[T1]{fontenc}
\usepackage{url}
\usepackage{booktabs}
\usepackage{amsfonts}
\usepackage{amsmath}
\usepackage{amssymb}
\usepackage{graphicx}
\usepackage{caption}
\usepackage{nicefrac}
\usepackage{microtype}
\usepackage{xcolor}
\usepackage{hyperref}

\title{RAG-RST: Retrieved Exemplars as In-Context Discourse Schema for RST Parsing}

\author{%
  Mohammadreza Sediqin \\
  Illinois Institute of Technology \\
  New York, USA \\
  \texttt{msediqin@hawk.illinoistech.edu}
  \And
  Shlomo Engelson Argamon \\
  Touro University \\
  New York, USA \\
  \texttt{shlomo.argamon@touro.edu}
}
\begin{document}

\maketitle
\begin{abstract}
In Rhetorical Structure Theory (RST), each edge of a discourse tree carries a
nuclearity label, marking which span holds the central content, and a relation
drawn from a closed inventory of functional categories. Neither label is a
surface property: the same two clauses can be Elaboration or Background
depending on rhetorical intent, and the boundary between them is fixed by
annotation convention rather than by lexical cues. Predicting these labels
therefore requires a formal discourse schema, and exposure to discourse-rich
pretraining text does not supply it: given only a target EDU pair, a strong LLM
recovers nuclearity at 0.794 F1 and relation at 0.738 F1. We propose RAG-RST, a
retrieval-augmented framework that supplies this schema as in-context
demonstration rather than through parameter updates, raising nuclearity to 0.846
F1 and relation to 0.788 F1 with no training. Two configurations are compared: a
precomputed strategy that labels all EDU pairs before parsing, and a dynamic
strategy that retrieves at each reduce step, so that the exemplars reaching the
generator depend on the reductions the parser has already committed to. Both
outperform existing RST parsers across span, nuclearity, relation, and full tree
metrics, with the dynamic variant consistently strongest. Two properties of the
retrieved schema emerge: its utility is bounded in number, saturating and then
declining beyond fifty exemplars, and it depends on structural position. Both
indicate that a formal linguistic schema can be supplied to a foundation model
as a demonstration rather than trained into its parameters.
\end{abstract}

\section{Introduction}

Discourse parsing aims to uncover the logical and coherence relations between text units in a structured and hierarchical manner. Rhetorical Structure Theory (RST) \citep{mann1988rhetorical} represents a text as a tree whose leaves are Elementary Discourse Units (EDUs) and whose internal nodes are composite spans formed by grouping them. Parsing combines adjacent EDUs into higher-level spans and labels each connection with nuclearity and relation. Neither label is a surface property of the text: nuclearity marks which span carries the central content, and the relation is drawn from a closed inventory of functional categories fixed by annotation convention rather than by lexical cues \citep{carlson2002rst}. The same two clauses can be Elaboration or Background depending on rhetorical intent, so predicting these labels requires a model to acquire a formal discourse schema rather than merely perform a task. Accurate labels matter downstream, in summarization, information extraction, and discourse-aware representation learning \citep{marcu2000theory, zhang2020top, sediqin2025rst}. As foundation models are increasingly deployed as agents, whose recorded outputs must be verified rather than merely scored \citep{sediqin2026trustlayeragentevaluation}, explicit discourse structure offers one way to make the reasoning in their generated text inspectable.

Methods for RST parsing range from top-down and bottom-up parsers that recursively divide or merge spans, to transition-based shift-reduce parsers that build the tree incrementally through actions managing and merging spans \citep{li2021neural}, to neural enhancements using contextualized span representations and deep encoders \citep{nguyen2021rst}. What these approaches share is that the annotation convention is acquired through training. Yet large pretrained models exposed to vast quantities of discourse-rich text still recover the RST schema poorly: given only a target EDU pair, a strong LLM reaches 0.794 F1 on nuclearity and 0.738 F1 on relation in our experiments. Scale over discourse-rich text is therefore not sufficient. The discourse phenomena are present in pretraining data; the annotation convention that carves them into these particular categories is not.

Retrieval-augmented generation (RAG) offers a way to supply that missing convention at inference time rather than through additional training. What form the supplied knowledge must take, however, remains open: labeled neighbors function as demonstrations of an annotation scheme, and their utility depends on how many are shown, how consistent they are, and where in the discourse structure the target pair sits \citep{mei2025survey}.

We introduce \textit{RAG-RST}, which integrates retrieval with a neural shift-reduce parser to improve nuclearity and relation prediction. Retrieval runs over the RST-annotated RST-GUM corpus \citep{zeldes2017gum}, which supplies gold labels; no RST-DT material is placed in the index. We additionally index sentence pairs from CNN/DailyMail \citep{hermann2015teaching} as an unlabeled discourse-rich corpus, used only to improve nearest-neighbor coverage and never to predict labels. To the best of our knowledge, this is the first work to supply the RST annotation schema through retrieval at inference time rather than through training, and the first to place retrieval inside the parsing loop, so that the exemplars shown to the generator depend on the reductions the parser has already committed to. Our contributions are as follows:
\begin{itemize}
    \item We propose RAG-RST, combining neural EDU encoders, FAISS retrieval, and large language models for label prediction, and compare two strategies: precomputed labeling, which generates labels for all EDU pairs before parsing, and dynamic label generation, which retrieves at each reduce step so that the exemplars reflect the reductions already committed to.
    \item We quantify how much of the RST annotation schema is unrecoverable from the EDU pair alone, separating a model's knowledge of discourse phenomena from its knowledge of the convention that categorizes them, and show that retrieved exemplars narrow this gap without any parameter update.
    \item We show that both strategies outperform existing RST parsers across all four standard evaluation metrics, with the dynamic approach consistently strongest.
    \item We scope our contribution to the label prediction stage: the shift/reduce action policy is adopted unmodified from prior work, and RAG-RST supplies nuclearity and relation labels to the reduce operation rather than altering how parser actions are selected.
\end{itemize}

\section{Related Work}
Discourse analysis based on RST has attracted considerable attention across NLP tasks \citep{sediqin2025esurfsimpleeffectiveedu}. Historically it has been approached as a parsing problem, using transition-based parsers that operate top-down or incrementally and chart parsers based on bottom-up dynamic programming \citep{soricut2003sentence, joty2015codra}, with later gains from statistical models that predict nuclearity and relation types between discourse units \citep{hernault2010hilda, liu2016implicit, wang2018toward}.

Among these, transition-based shift-reduce parsers have become a standard approach for RST tree construction, offering efficient incremental tree building through generic actions that merge spans and assign nuclearity and relation labels during parsing \citep{sagae2009analysis, li2021neural}. Neural enhancements, including contextualized span representations and deep encoders, significantly improve parsing performance and robustness \citep{nguyen2021rst}. Despite these advances, accurate label prediction remains a challenge in complex discourse structures \citep{maekawa2024can, zhang2025self}. The limitation is concentrated in the label prediction stage: nuclearity and relation assignment follows span segmentation and directly determines the quality of the constructed tree \citep{liu2020multilingual, zhang2020top}, and performance gaps are most evident where relations are ambiguous or span long distances. In every case the annotation convention is acquired during training, so the schema is fixed at deployment and cannot be revised without retraining.

Retrieval-augmented generation offers one route to supplying such knowledge at inference time \citep{lewis2020retrieval, sediqin2026sagegovernedartifactgeneration}. In discourse parsing, RAG lets a model condition nuclearity and relation prediction on relevant examples from annotated corpora. The closest prior work is \citet{liu2026disco}, which applies retrieval to discourse parsing but does so during pretraining, learning discourse-aware representations rather than retrieving at inference time. Our approach differs in two respects. First, we retrieve and generate at parse time, at the EDU pair level, so the system adapts to each new document without parameter updates. Second, prior work does not place retrieval inside the parsing loop. Our Dynamic RAG variant is the first in which the discourse structure already built determines which exemplars the generator sees, so that the same EDU pair retrieves different demonstrations depending on how earlier reductions grouped it. We instantiate this as \textit{RAG Enhanced RST Parsing (RAG-RST)}, evaluated on overall parsing performance and specifically on nuclearity and relation label prediction.

\section{RAG-Enhanced RST Discourse Parsing Framework}

We propose a two-mode framework that integrates Retrieval-Augmented Generation (RAG) with a shift-reduce RST discourse parser. The system operates in two configurations: (1) Precomputed Labeling outside the parser, and (2) RAG During Reduce action inside the parser. Both methods rely on a FAISS-based dense retriever and LLMs as the generation model, but differ in how they interact with the parser during tree construction. Figures~\ref{fig:pos1} and~\ref{fig:pos2} illustrate the workflows for the Precomputed Labeling and RAG During Reduce methods, respectively. These diagrams summarize the key components, including EDU segmentation with LACES \citep{sediqin2025laces}, transformer-based encoding, FAISS retrieval, LLM label generation, and integration with the shift-reduce parser.

\subsection{Method 1: Precomputed Labeling (Outside the Parser)}
\label{sec:precomputed}

\begin{figure}[t]
\centering
  \includegraphics[width=0.8\textwidth]{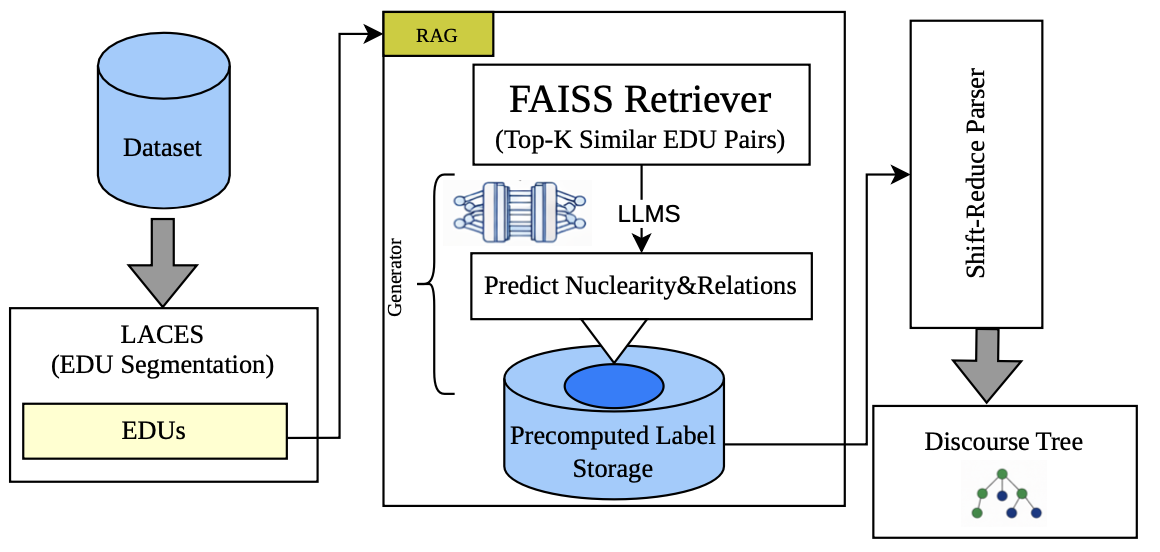} 
  \vspace{-3mm}
  \caption{Architecture of the Precomputed Labeling pipeline.}
  \label{fig:pos1}
\end{figure}

In this configuration, nuclearity and discourse relation labels are generated before the discourse parsing stage using a RAG pipeline. The predicted labels are precomputed and stored in an external lookup table. The shift reduce parser then uses these stored labels to guide discourse tree construction without invoking the retrieval or generation components during parsing (Figure~\ref{fig:pos1}).

\begin{itemize}

\item \textbf{Encoding Annotated EDU Pairs}: The annotated corpus provides EDU pairs $(e_i, e_j)$ together with their gold nuclearity and discourse relation labels. Each EDU is encoded using a transformer encoder as $\mathbf{h}_i = f_{\text{enc}}(e_i)$ and $\mathbf{h}_j = f_{\text{enc}}(e_j)$. A pair level representation is constructed by combining the two embeddings:
\begin{align}
\label{eq:pair}
    \mathbf{v}_{ij} =
[\mathbf{h}_i \,\Vert\, \mathbf{h}_j \,\Vert\, |\mathbf{h}_i - \mathbf{h}_j|]
\end{align}
This representation captures both the semantic content of the EDUs and their relational interaction.

\item \textbf{FAISS Dense Retriever}: All annotated EDU pair vectors $\mathbf{v}_{ij}$ together with their corresponding labels are stored in a FAISS index to enable efficient similarity retrieval. For a new EDU pair representation $\mathbf{v}_{ij}$, the retriever identifies the top $K$ most similar annotated pairs:
\begin{align}
\label{eq:faiss}
\mathcal{N}_K(\mathbf{v}_{ij}) =
\text{FAISS.topK}(\mathbf{v}_{ij})    
\end{align}

\item \textbf{EDU Segmentation for New Input}: For a new document, EDUs are first obtained through segmentation. We leverage LACES model, a lightweight yet effective EDU segmenter that predicts EDU boundaries using a sliding nine token window. The resulting EDUs are paired and encoded using the same transformer encoder to produce the representation $\mathbf{v}_{ij}$ used for FAISS retrieval.

\item \textbf{LLM Generator for Nuclearity and Relation Prediction}: The retrieved annotated examples $\mathcal{N}_K(\mathbf{v}_{ij})$ together with the target EDU pair $(e_i, e_j)$ are provided to an LLM, which generates the predicted discourse relations:
\begin{align}
\label{eq:gen}
    (\hat{n}_{ij}, \hat{r}_{ij}) =
g_{\text{LLM}}\bigl(e_i, e_j, \mathcal{N}_K(\mathbf{v}_{ij})\bigr)
\end{align}
Here, $\mathcal{N}_K(\mathbf{v}_{ij})$ denotes the set of retrieved annotated EDU pairs and their gold labels; in practice, the target EDU pair and the retrieved examples are concatenated into a structured input for the LLM, which generates the nuclearity and relation labels as structured text outputs using greedy decoding.

\item \textbf{Shift-Reduce Parsing with Precomputed Labels}: The predicted labels are stored prior to parsing in an external lookup table $\mathcal{L} = \{(i, j, \hat{n}_{ij}, \hat{r}_{ij})\}$, and a shift reduce discourse parser then constructs the discourse tree for each document. The parser maintains a stack of partially built spans and a buffer containing the remaining EDUs. At each step, the parser state is represented as:
\begin{align}
\label{eq:state}
 \mathbf{p}_t = f_{\text{state}}(s_t, b_t)   
\end{align}
where $s_t$ denotes the stack state and $b_t$ denotes the remaining EDUs in the buffer. The parser performs two primary actions:

\begin{itemize}
\item \textbf{Shift:} The next EDU from the buffer is pushed onto the stack.
\item \textbf{Reduce:} The parser combines the top spans on the stack into a larger discourse unit. During this step, the parser retrieves the corresponding nuclearity and relation labels from the precomputed lookup table $\mathcal{L}$ to guide the reduction.
\end{itemize}

The parser iteratively applies shift and reduce operations until a complete discourse tree is constructed for the document. By generating discourse labels prior to parsing, this approach separates label prediction from tree construction while allowing the parser to benefit from enriched relational information produced by the retrieval-augmented generation module.

\end{itemize}

\subsection{Method 2: RAG During Reduce Action (Inside the Parser)}
\label{sec:dynamic}

\begin{figure}[t]
\centering
  \includegraphics[width=0.8\textwidth]{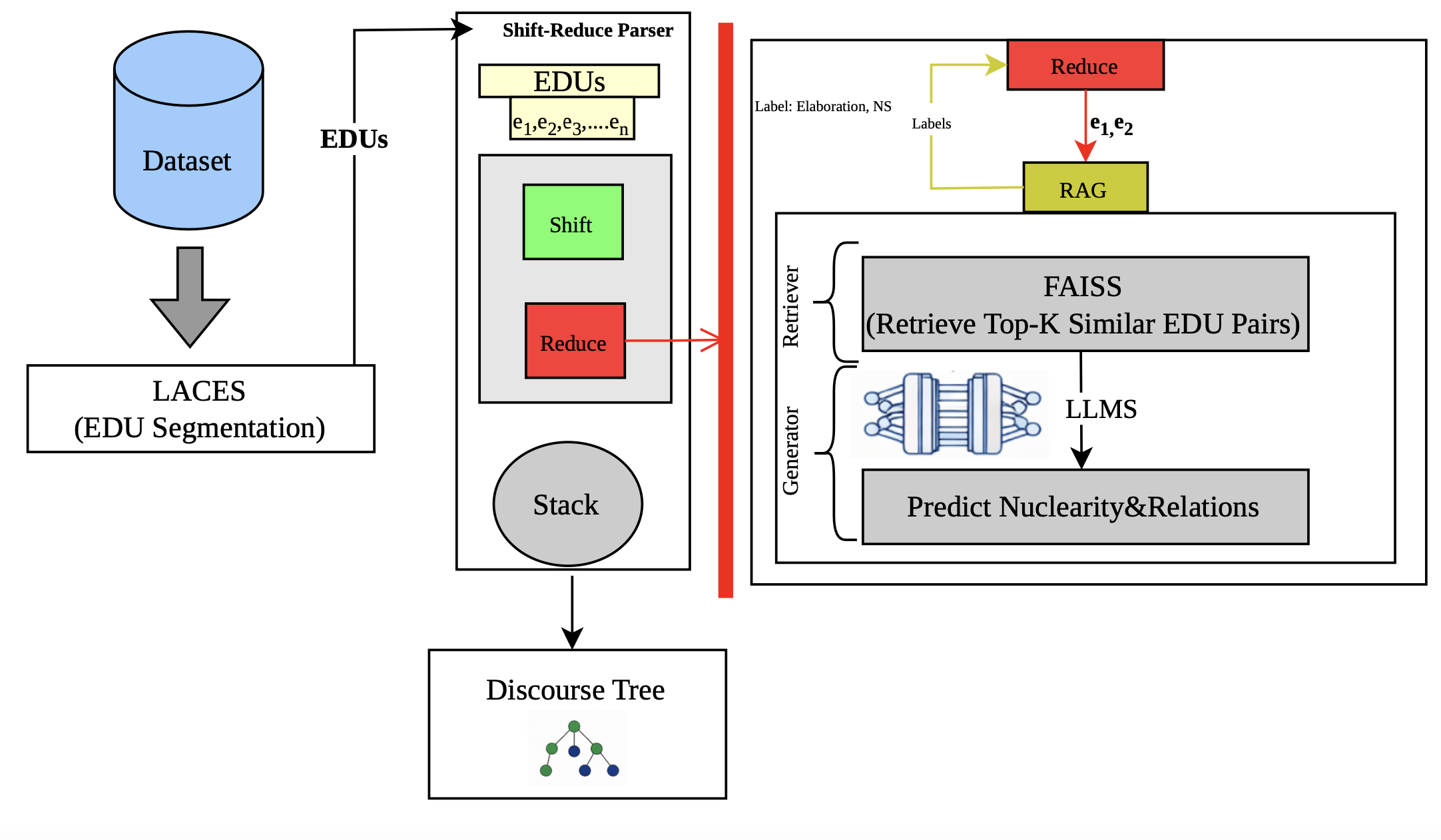} 
  \vspace{-3mm}
  \caption{Architecture of the Dynamic RAG During Reduce Action.}
  \label{fig:pos2}
\end{figure}

This method integrates RAG dynamically within the shift-reduce parser, applying generator predictions at every reduce action (Figure~\ref{fig:pos2}). Segmentation, pair encoding, and label generation follow Section~\ref{sec:precomputed}; what changes is when retrieval is invoked and what the query represents.

\begin{itemize}

\item \textbf{Shift Reduce Parser State}: The parser maintains a stack of partially constructed discourse units and a buffer of remaining EDUs obtained from LACES segmentation. At each step, the parser chooses between Shift and Reduce actions, with the parser state at time $t$ represented by Eq.~\ref{eq:state}.

\item \textbf{Parser Policy and Span Encoding}: The choice between shift and reduce at each parser state is made by a standard neural action classifier, which we adopt without modification. When the top of the stack contains a composite span (an already reduced subtree) rather than a single EDU, we encode the span as the mean pooled representation of all EDUs it contains. This pooled vector replaces $\mathbf{h}_i$ or $\mathbf{h}_j$ in the pair representation $\mathbf{v}_{ij}$ used to query FAISS. The retrieval query is not a function of the full parser state in Eq.~\ref{eq:state}: $\mathbf{v}_{ij}$ is built from the two stack-top spans only, excluding the buffer and the deeper stack. When both operands are single EDUs, the Dynamic query is identical to the Precomputed one by construction, so the two configurations differ only at reduce actions where at least one operand is a composite span. At those reductions the pooled representation of a composite span reflects the grouping the parser has already committed to, so the same EDU pair yields a different query depending on how earlier reductions arranged it. These are also the reductions where the relation holds between composite spans rather than between adjacent clauses, and where the relation between two spans is not in general the relation holding between their leftmost leaves. This is the restricted sense in which retrieval here is conditioned on parsing history.

\item \textbf{FAISS Retrieval and Label Prediction During Reduce}: When a reduce action is triggered, the parser selects the top EDU pair $(e_i,e_j)$ from the stack and constructs a pair representation as in Eq.~\ref{eq:pair}, which queries a FAISS index built from annotated EDU pairs in the RST-GUM dataset, augmented with unlabeled EDU pairs from CNN/DailyMail as described in Section~\ref{sec:retrieval_setup}. The retriever returns the top $K$ most similar annotated pairs (Eq.~\ref{eq:faiss}), which are passed with the target pair to the generator (Eq.~\ref{eq:gen}). The predicted labels are applied immediately, and the labeled span is pushed back onto the stack. This process continues until a complete discourse tree is constructed for the document. \item \textbf{Dynamic RAG Integrated Parsing}: The parser iteratively performs shift and reduce actions. At each reduce step the retrieval and generation modules are invoked on the current stack-top spans, the predicted nuclearity and relation labels are applied immediately, and the labeled span is pushed back onto the stack. Label prediction is therefore performed at decision time rather than read from a table, so each prediction is made in the structural context the parser has reached.

\end{itemize}

\section{Experiments}
In this section, we describe our datasets, baselines, experimental setup, and results.

\subsection{Experimental Setup}

\subsubsection{Dataset}

We follow standard evaluation practices and use two RST benchmark datasets for training and evaluation:

\begin{itemize}
    \item RST Discourse Treebank (RST-DT) \citep{carlson2002rst}, the primary benchmark for RST parsing, containing 347 training and 38 test documents with full discourse tree annotations.
    \item Georgetown University Multilayer Corpus (RST-GUM) \citep{zeldes2017gum}, a multi genre discourse corpus used as the external annotated source for retrieval in our RAG framework. It provides diverse discourse structures that improve retrieval coverage.
\end{itemize}

In addition to these RST annotated corpora, we incorporate the CNN/Daily Mail dataset \citep{hermann2015teaching} as an auxiliary unlabeled discourse rich corpus, used only to enrich the retrieval space with additional semantic discourse patterns. It is not used for training, evaluation, or supervision of RST labels, so all evaluation is performed strictly on standard RST benchmarks while still benefiting from broader discourse like textual patterns during retrieval.

\subsubsection{Baselines for RST Parsing}
We compare against recent state of the art RST parsers covering the major modeling paradigms, all widely used as standard baselines and therefore directly comparable with prior work. \citet{feng2014linear} propose a CRF based parser, and \citet{hayashi2016empirical} reimplement HILDA with SVM classification for bottom up parsing. \citet{maekawa2024can} represent LLM based methods, exploring prompting and fine tuning for end to end RST tree generation. This system is also the closest supervised comparison to ours: it acquires the RST annotation schema through parameter updates on RST-annotated data, whereas RAG-RST supplies the same schema through retrieved exemplars at inference time with no updates. Table~\ref{tab:parser_results} therefore contrasts the two routes at the level of full parsing performance, though not at the label prediction level of Tables~\ref{tab:retrieval_ablation} and~\ref{tab:generator_comparison}. \citet{zhang2021adversarial} propose a neural top down parser with a generator and discriminator framework for global structure consistency, and \citet{nguyen2021rst} a sequence to sequence transition based parser built on a pointer network. \citet{zhang2020top} employ syntax aware graph attention networks for joint structure and relation modeling, while \citet{koto2021top} cast parsing as sequence labeling with a transformer based top down parser and a dynamic oracle. \citet{yu2022rst} add EDU level pretraining to improve discourse representations, and \citet{sediqin2025laces} improve downstream parsing through more accurate EDU segmentation combined with a transition based parser.

\subsubsection{Retrieval and Generator Setup}
\label{sec:retrieval_setup}

Both configurations share a unified retrieval and generation pipeline. Documents from RST-DT are segmented into EDUs using the LACES segmenter, and each resulting EDU pair is encoded with \texttt{sentence-transformers/all-MiniLM-L6-v2} to produce the query embedding for retrieval. We use MiniLM for its low encoding cost, which matters because the Dynamic variant re-encodes stack-top spans at every reduce action; stronger retrievers would likely improve neighbor quality further. We build a FAISS index over annotated EDU pairs from RST-GUM, augmented with unlabeled but discourse rich EDU pairs from CNN/DailyMail for broader semantic coverage. The top $K$ most similar examples are retrieved by cosine similarity, with an empirically selected threshold of 0.6: lower values admit weakly related or noisy pairs that degrade label prediction, while higher ones reduce retrieval coverage and limit diversity in the contextual examples available to the generator. Both $K$ and the similarity threshold were selected on a development split drawn from the 347 RST-DT training documents, on which all label prediction results are also reported. The 38 test documents are used only for the final parsing evaluation in Table~\ref{tab:parser_results}.

All generator models operate in a text-to-text setting with greedy decoding at inference. The input concatenates the target EDU pair with the retrieved labeled neighbors, and the model generates the nuclearity label followed by an RST relation from the inventory of \citet{carlson2002rst}. The three nuclearity labels indicate which span carries the central content: N-S marks the left span as nucleus and the right as satellite, S-N marks the right span as nucleus, and N-N indicates two nuclei of equal importance. CNN/DailyMail neighbors, which lack RST annotations, are excluded from the labeled context and contribute only to embedding space coverage. Experiments run on an Ubuntu 20.04.4 server with a single NVIDIA A100 80GB GPU, and reported results are means over three runs with different random seeds, following standard practice in prior RST parsing work \citep{maekawa2024can, nguyen2021rst, yu2022rst}; variance was small ($\pm 0.005$ F1 or less on all metrics), indicating stable performance.

No RST-DT material is placed in the retrieval index at any stage. RST-GUM and RST-DT are drawn from disjoint sources and share no documents. The CNN/DailyMail index was screened against the 38 RST-DT test documents for near-duplicate passages before indexing, and matching passages were excluded \citet{sediqin2026gvdgovernedversioningdeduplication}  Retrieved neighbors therefore originate only from documents the evaluation never sees. CNN/DailyMail entries carry no RST annotation and are excluded from the labeled context supplied to the generator in every configuration, so they cannot transmit label information to the evaluation.

\subsubsection{Parser Configurations}
\label{sec:parser_configs}

Both configurations use the pipeline described above, with LLaMA-3 as the generator and $K=50$ retrieved neighbors, and differ only in when retrieval is invoked. In the Precomputed setting (Section~\ref{sec:precomputed}), labels are generated for every adjacent EDU pair in a document before parsing begins and stored in a lookup table, which the shift reduce parser consults at each reduce action. In the Dynamic setting (Section~\ref{sec:dynamic}), retrieval and generation are invoked at each reduce action using the current stack-top spans, so that label prediction is performed at decision time while the tree is being constructed.

\subsection{Results}

\subsubsection{Schema Recoverability and the Effect of Retrieval}
We first ask how much of the RST label schema the generator recovers without retrieval. Three configurations isolate the contribution of retrieval and the role of CNN/DailyMail augmentation: LLaMA-3 with no retrieval, with retrieval over RST-GUM only, and with retrieval over RST-GUM augmented by CNN/DailyMail. All three share the same generator and the same LACES segmentation, so the only difference is what the generator receives: the target EDU pair alone, or that pair together with the top 50 nearest neighbors and their gold labels. Table~\ref{tab:retrieval_ablation} reports nuclearity and relation F1 at $K=50$, the retrieval size selected in the sweep reported below.

\begin{table}[htbp]
\centering
\caption{Effect of retrieval on label prediction quality ($K=50$, LLaMA-3), evaluated on the RST-DT development split. Deltas are measured against the no retrieval baseline.}
\label{tab:retrieval_ablation}
\begin{tabular}{lcccc}
\toprule
\textbf{Configuration} & \multicolumn{2}{c}{\textbf{Nuclearity}} & 
\multicolumn{2}{c}{\textbf{Relation}} \\
\cmidrule(lr){2-3} \cmidrule(lr){4-5}
 & F1 & $\Delta$ & F1 & $\Delta$ \\
\midrule
LLaMA-3, no retrieval & 0.794 & & 0.738 & \\
RST-GUM only          & 0.829 & +0.035 & 0.768 & +0.030 \\
RST-GUM + CNN/DM      & \textbf{0.846} & +0.052 & \textbf{0.788} & +0.050 \\
\bottomrule
\end{tabular}
\end{table}

Retrieval accounts for the majority of the label prediction gain. Without it, nuclearity falls from 0.846 to 0.794 and relation from 0.788 to 0.738, so the retrieved labeled neighbors supply signal the generator cannot recover from the EDU pair alone. The discourse phenomena underlying these labels are pervasive in pretraining text, which locates the gap in the annotation convention rather than in the linguistic content it categorizes: what the model lacks is not the ability to see that one span elaborates another, but the convention that decides where Elaboration ends and Background begins.

Adding CNN/DailyMail yields a further +0.017 nuclearity and +0.020 relation over RST-GUM alone. These examples carry no RST annotation and are excluded from the labeled context, so the gain arises from their presence in the index changing which labeled neighbors reach the generator and how many. Its direction matches the sweep in Table~\ref{tab:nuclearity_relation_faiss}, where performance peaks at $K=50$ and declines beyond it, both pointing to a bounded capacity to absorb schema from context rather than a monotonic benefit from more retrieved material. Isolating the mechanism would require varying the composition of the retrieved set while holding the number of labeled exemplars fixed.

\subsubsection{Retrieval Size and Generator Results}

If retrieved neighbors function as demonstrations of the annotation scheme rather than as additional evidence about the target pair, their utility should be bounded rather than monotonic in the number supplied. We evaluate $K \in \{20, 40, 50, 60, 70\}$ to determine the optimal retrieval size. As shown in Table~\ref{tab:nuclearity_relation_faiss}, LLaMA-3 (8B base), LLaMA-2 (7B base), and T5 (base) all benefit from increasing $K$ up to 50, with consistent improvements across precision, recall, and F1 score for both nuclearity and relation prediction tasks. In particular, $K=50$ achieves the best performance across all models. LLaMA-3 reaches the highest F1 scores of 0.846 (nuclearity) and 0.788 (relation), followed closely by LLaMA-2 with 0.841 (nuclearity) and 0.780 (relation), and T5 with 0.779 (nuclearity) and 0.735 (relation). Beyond $K=50$, performance degrades across all models, consistent with diminishing utility from lower-ranked exemplars. The saturation and subsequent decline is what the demonstration account predicts and is not what additional evidence about the target pair would produce, since more evidence should not hurt. Top-50 retrieval is therefore selected on the development split and fixed for all subsequent experiments, including the final evaluation on the held-out test set.

\begin{table}[htbp]
\centering
\caption{Nuclearity and relation F1 across Top-K values using FAISS retrieval, evaluated on the RST-DT development split. Full precision and recall figures are reported in Appendix~\ref{app:full}.}
\label{tab:nuclearity_relation_faiss}
\small
\begin{tabular}{ccccccc}
\toprule
\textbf{K} 
& \multicolumn{2}{c}{\textbf{LLaMA-3}} 
& \multicolumn{2}{c}{\textbf{LLaMA-2}} 
& \multicolumn{2}{c}{\textbf{T5}} \\
\cmidrule(lr){2-3} \cmidrule(lr){4-5} \cmidrule(lr){6-7}
& Nuc & Rel & Nuc & Rel & Nuc & Rel \\
\midrule
20 & 0.833 & 0.728 & 0.825 & 0.718 & 0.768 & 0.691 \\
40 & 0.843 & 0.761 & 0.835 & 0.752 & 0.774 & 0.715 \\
\textbf{50} & \textbf{0.846} & \textbf{0.788} & \textbf{0.841} & \textbf{0.780} & \textbf{0.779} & \textbf{0.735} \\
60 & 0.844 & 0.782 & 0.838 & 0.761 & 0.776 & 0.718 \\
70 & 0.840 & 0.771 & 0.833 & 0.747 & 0.773 & 0.706 \\
\bottomrule
\end{tabular}
\end{table}

For generation, we evaluate six architectures: LLaMA-3 (8B base), LLaMA-2 (7B base), Mistral-7B, T5 (base), BART (large), and RoBERTa (large), under the same retrieval setting with Top-50 FAISS retrieval. As Table~\ref{tab:generator_comparison} demonstrates, LLaMA-3 achieves the best overall performance, obtaining the highest F1 scores for nuclearity (0.846), relation (0.788), and combined metrics (0.817). LLaMA-2 and Mistral-7B perform comparably and both substantially outperform encoder only and sequence to sequence models, confirming that autoregressive decoder based LLMs are better suited for leveraging retrieved contextual examples in discourse label prediction. Based on these results, we select LLaMA-3 as the generator model for both Precomputed and Dynamic RAG frameworks.

\begin{table}[htbp]
\centering
\caption{Generator model comparison (Top-50 EDUs) on the RST-DT development split. Combined scores are computed jointly over nuclearity and relation predictions.}
\label{tab:generator_comparison}
\small
\begin{tabular}{lccccccccc}
\toprule
\textbf{Model} & \multicolumn{3}{c}{\textbf{Nuclearity}} & 
\multicolumn{3}{c}{\textbf{Relation}} & 
\multicolumn{3}{c}{\textbf{Combined}} \\
\cmidrule(lr){2-4} \cmidrule(lr){5-7} \cmidrule(lr){8-10}
 & Prec & Rec & F1 & Prec & Rec & F1 & Prec & Rec & F1 \\
\midrule
\textbf{LLaMA-3}  & 0.851 & 0.842 & \textbf{0.846} & 0.796 & 0.781 & \textbf{0.788} & 0.822 & 0.809 & \textbf{0.817} \\
LLaMA-2  & 0.845 & 0.835 & 0.841 & 0.788 & 0.772 & 0.780 & 0.815 & 0.800 & 0.807 \\
Mistral-7B        & 0.843 & 0.833 & 0.838 & 0.785 & 0.769 & 0.777 & 0.812 & 0.798 & 0.805 \\
T5                & 0.780 & 0.778 & 0.779 & 0.742 & 0.728 & 0.735 & 0.761 & 0.753 & 0.757 \\
BART              & 0.798 & 0.790 & 0.794 & 0.760 & 0.748 & 0.754 & 0.779 & 0.769 & 0.774 \\
RoBERTa           & 0.763 & 0.752 & 0.757 & 0.730 & 0.718 & 0.724 & 0.747 & 0.740 & 0.743 \\
\bottomrule
\end{tabular}
\end{table}

\subsubsection{RST Parsing Results}

We now examine whether these label level effects carry over to tree construction on the test set, where labeling errors compound across reductions. Table~\ref{tab:parser_results} compares both proposed configurations with existing baselines, using LLaMA-3 as the generator throughout since Table~\ref{tab:generator_comparison} established it as the strongest. Following \citet{morey2017much}, \citet{shahmohammadi2024discourse}, \citet{yu2022rst}, and \citet{zhang2021adversarial}, we report micro averaged F1 across four metrics: \textbf{S}pan (tree structure without labels), \textbf{N}uclearity (structure with nuclearity labels), \textbf{R}elation (structure with relation labels), and \textbf{F}ull (structure with both). All scores use the original RST-Parseval variant, and baseline figures are taken from published results on the same RST-DT split.

\begin{table}[htbp]
\centering
\caption{Comparison of RST parser performance across Precomputed RAG, Dynamic RAG, and baseline systems. All metrics are micro averaged F1 scores.}
\label{tab:parser_results}
\small
\begin{tabular}{lcccc}
\toprule
\textbf{Model} & \textbf{S} & \textbf{N} & \textbf{R} & \textbf{F} \\
\midrule
\citet{sediqin2025laces} & 0.879 & 0.762 & 0.641 & 0.634 \\
\citet{feng2014linear} & 0.686 & 0.559 & 0.458 & 0.446 \\
\citet{hayashi2016empirical} & 0.651 & 0.546 & 0.447 & 0.441 \\ 
\citet{maekawa2024can} & 0.798 & 0.704 & 0.600 & 0.581 \\
\citet{zhang2021adversarial} & 0.763 & 0.655 & 0.556 & 0.538 \\
\citet{nguyen2021rst} & 0.743 & 0.643 & 0.516 & 0.502 \\
\citet{koto2021top} & 0.731 & 0.623 & 0.515 & 0.503 \\
\citet{yu2022rst} & 0.714 & 0.603 & 0.492 & 0.481 \\
\citet{zhang2020top} & 0.672 & 0.555 & 0.453 & 0.443 \\
\midrule
RAG + LLaMA-3 (Precomputed) & 0.885 & 0.782 & 0.694 & 0.663 \\
\textbf{RAG + LLaMA-3 (Dynamic)} & \textbf{0.892} & \textbf{0.791} & \textbf{0.711} & \textbf{0.674} \\
\bottomrule
\end{tabular}
\end{table}

Both configurations outperform every baseline on all four metrics. Precomputed labeling reaches 0.885 span, 0.782 nuclearity, 0.694 relation, and 0.663 full tree, which establishes that retrieval augmented label generation is effective even when performed entirely before parsing. Dynamic RAG improves further to 0.892, 0.791, 0.711, and 0.674. Since the two variants share index, encoder, generator, and segmenter, and differ only in whether the retrieval query reflects the reductions the parser has already committed to, the gap between them isolates the effect of structural conditioning on exemplar selection. The two ablations therefore decompose the gain: Table~\ref{tab:retrieval_ablation} attributes most of the label prediction improvement to retrieval itself, while the Precomputed to Dynamic difference measures what parser integration adds on top. Gains of this size are substantial in RST parsing, where structural and semantic complexity limits achievable improvement \citep{morey2017much}.

The advantage of the Dynamic variant follows from what its query represents. Precomputed retrieval operates over statically paired EDUs before parsing begins, so the query captures only the local semantic content of the pair in isolation. Dynamic retrieval instead runs at each reduce step, where the query reflects the pair as it stands within the partially constructed tree and is therefore conditioned on the structural decisions already made. Which exemplars best demonstrate the annotation scheme thus depends on structural position, since the relation between two composite spans is not in general the relation holding between their leftmost leaves. Retrieving at the point of decision accordingly produces more coherent label sequences across the tree, particularly for complex or long distance relations that local context alone does not resolve.

\section{Limitations}
Our results are established on English discourse, and extending the framework to other languages remains an important direction. Since the approach requires no parameter updates and depends only on retrieved exemplars, it transfers naturally to any language with an annotated RST corpus. Our sweep covers retrieval size and generator architecture, with MiniLM chosen for encoding cost and greedy decoding following standard practice, so stronger retrievers would be expected to improve neighbor quality and the reported figures are conservative in that respect. We evaluate retrieval through its effect on label prediction rather than intrinsically, and intrinsic measures such as neighbor label agreement would characterize the contribution of individual neighbors further.

\section{Conclusion and Future Work}
In this paper we propose RAG-RST, a retrieval-augmented framework that supplies the RST annotation schema to a foundation model as in-context demonstration rather than through parameter updates, in two configurations: Precomputed labeling before parsing and Dynamic retrieval at each reduce step. Given only an EDU pair, a strong LLM recovers the annotation convention poorly, yet fifty labeled neighbors recover much of that distance with no training. The discourse phenomena these labels describe are pervasive in pretraining text; what retrieval supplies is the convention that carves them into categories, which is why exemplars help where scale alone does not. Both configurations outperform existing parsers across span, nuclearity, relation, and full tree metrics, with retrieval at the point of decision consistently stronger than retrieval in advance. Two properties of the retrieved schema emerge: its utility is bounded in number, saturating and then declining beyond fifty exemplars, and it depends on structural position, since the relation between two composite spans is not in general the relation holding between their leftmost leaves. Both point to in-context exemplars functioning as demonstrations of an annotation convention rather than as additional evidence about the target pair. Future work will explore stronger retrieval models and larger or instruction-tuned generators, integrate the framework into other RST parsers, evaluate it across languages, and investigate downstream tasks such as summarization.

\bibliographystyle{plainnat}
\bibliography{reference}

\clearpage
\appendix

\section{Full Retrieval Results}
\label{app:full}

Table~\ref{tab:faiss_full} reports precision, recall, and F1 for all three generator models across the full range of $K$ values evaluated, extending the F1 figures summarized in Table~\ref{tab:nuclearity_relation_faiss}.

\begin{table}[htbp]
\centering
\caption{Nuclearity and relation prediction performance across Top-K values using FAISS retrieval on the RST-DT development split, with full precision and recall.}
\label{tab:faiss_full}
\scriptsize
\setlength{\tabcolsep}{3pt}
\begin{tabular}{ccccccccccccccccccc}
\toprule
\textbf{K} 
& \multicolumn{6}{c}{\textbf{LLaMA-3}} 
& \multicolumn{6}{c}{\textbf{LLaMA-2}} 
& \multicolumn{6}{c}{\textbf{T5}} \\
\cmidrule(lr){2-7} \cmidrule(lr){8-13} \cmidrule(lr){14-19}
& \multicolumn{3}{c}{Nuclearity} & \multicolumn{3}{c}{Relation} 
& \multicolumn{3}{c}{Nuclearity} & \multicolumn{3}{c}{Relation}
& \multicolumn{3}{c}{Nuclearity} & \multicolumn{3}{c}{Relation} \\
\cmidrule(lr){2-4} \cmidrule(lr){5-7} \cmidrule(lr){8-10} \cmidrule(lr){11-13} \cmidrule(lr){14-16} \cmidrule(lr){17-19}
& P & R & F1 & P & R & F1 & P & R & F1 & P & R & F1 & P & R & F1 & P & R & F1 \\
\midrule
20 & 0.838 & 0.829 & 0.833 & 0.741 & 0.715 & 0.728 & 0.830 & 0.820 & 0.825 & 0.732 & 0.705 & 0.718 & 0.768 & 0.768 & 0.768 & 0.702 & 0.680 & 0.691 \\
40 & 0.848 & 0.839 & 0.843 & 0.771 & 0.751 & 0.761 & 0.840 & 0.831 & 0.835 & 0.762 & 0.742 & 0.752 & 0.772 & 0.776 & 0.774 & 0.724 & 0.707 & 0.715 \\
50 & 0.851 & 0.842 & 0.846 & 0.796 & 0.781 & 0.788 & 0.845 & 0.835 & 0.841 & 0.788 & 0.772 & 0.780 & 0.780 & 0.778 & 0.779 & 0.742 & 0.728 & 0.735 \\
60 & 0.848 & 0.840 & 0.844 & 0.792 & 0.773 & 0.782 & 0.844 & 0.832 & 0.838 & 0.770 & 0.752 & 0.761 & 0.777 & 0.775 & 0.776 & 0.727 & 0.710 & 0.718 \\
70 & 0.846 & 0.835 & 0.840 & 0.781 & 0.762 & 0.771 & 0.839 & 0.828 & 0.833 & 0.757 & 0.738 & 0.747 & 0.774 & 0.773 & 0.773 & 0.715 & 0.698 & 0.706 \\
\bottomrule
\end{tabular}
\end{table}

\end{document}